\documentclass[a4paper, 10pt, conference]{ieeeconf}
\IEEEoverridecommandlockouts                              % This command is only
\usepackage{epsfig, subfigure} % for postscript graphics files
\usepackage{amsmath} % assumes amsmath package installed
\usepackage{amssymb}  % assumes amsmath package installed
\newcommand \rep{{\bf\sffamily{Replicator/Symbrion}}}
\pdfoutput=1
\newif\ifpdf\ifx\pdfoutput\undefined\pdffalse\else\pdfoutput=1\pdftrue\fi

\title{\LARGE \bf Multi-Robot Organisms: State of the Art}

\author{{Serge Kernbach$^1$\footnote{Contact author: korniesi@ipvs.uni-stuttgart.de}, Oliver Scholz$^2$,
Kanako Harada$^3$, Sergej Popesku$^1$, Jens Liedke$^4$, Humza Raja$^2$,}\\
{Wenguo Liu$^5$, Fabio Caparrelli$^6$, Jaouhar Jemai$^7$, Jiri Havlik$^8$, Eugen Meister$^1$, and Paul Levi$^1$}\\[3mm]
\small
$^1$Institute of Parallel and Distributed Systems, University of Stuttgart, Germany,\\
$^2$Fraunhofer Institute for Biomedical Engineering, Sankt Ingbert, Germany,\\
$^3$Center for Applied Research in Micro and Nano Engineering , Scuola Superiore Sant'Anna, Italy,\\
$^4$Institute for Process Control and Robotics, Universit\"at Karlsruhe (TH), Germany,\\
$^5$Bristol Robotics Laboratory (BRL), UWE Bristol, UK,\\
$^6$Materials and Engineering Research Institute, Sheffield Hallam University, UK,\\
$^7$Ubisense AG, Munich, Germany, $^8$IMA, s.r.o., Prague, Czech Republic.\\
\thanks{Contact author: korniesi@ipvs.uni-stuttgart.de. appeared in ICRA2010, workshop on ``Modular Robots: State of the Art", pp.1-10, Anchorage, 2010.} }
\normalsize
\begin{document}

\maketitle
\thispagestyle{empty}
\pagestyle{empty}

%%%%%%%%%%%%%%%%%%%%%%%%%%%%%%%%%%%%%%%%%%%%%%%%%%%%%%%%%%%%%%%%%%%%%%%%%%%%%%%%
\begin{abstract}
This paper represents the state of the art development on the field of artificial multi-robot organisms. It briefly considers mechatronic development, sensor and computational equipment, software framework and introduces one of the Grand Challenges for swarm and reconfigurable robotics.
\end{abstract}
%%%%%%%%%%%%%%%%%%%%%%%%%%%%%%%%%%%%%%%%%%%%%%%%%%%%%%%%%%%%%%%%%%%%%%%%%%%%%%%%

\section{Introduction}

Appearance of multicellular structures is related to one of the greatest moments in the history of life~\cite{Alberts08}. The rise of multicellular from unicellular is a huge evolutionary step, however we do not exactly know how multicellular organisms appear and which mechanisms take part in this phenomenon. We know multicellular organisms are self-adaptive, self-regulative and self-developing, however we do not know its evolutionary origin and developmental organization. The great vision, which consolidates many interdisciplinary researchers, is a vision of self-adaptive, self-regulative and self-developing robots that reflect multicellularity in nature -- a vision of artificial robot organisms~\cite{Levi10}. Like multicellular beings, these artificial organisms consist of many small cell-modules, which can act as one structure and can exchange information and energy within this structure. Moreover, these structures can repair themselves and undergo evolutionary development from simple to complex organisms~\cite{Kernbach08b}.

Technological exploitation of multicellularity provides different practical advantages not only for advanced robotics, but also for autonomous and adaptive systems in general. Three most important advantages are extended reliability, advanced adaptivity and self-evolving properties. Reliability in general context is related to the ability of a system to work durably in different hostile or unexpected circumstances. Artificial organisms can self-disassemble, the destroyed cell-modules should be removed, and then an organism self-assembles again. Capabilities of basic robot modules for autonomous self-assembling and for dynamic change of functionality are key points of the extended reliability. Adaptivity is another key feature of advanced autonomous systems and indicates an ability of a system to cope with a changing environment. Multicellularity introduces a new component into adaptive processes -- morphogenesis -- the self-development of structure, functionality and behavior during a life cycle of the organism.
Both reliability and adaptivity mean a high developmental plasticity, where an organism can dynamically change itself, modify its own structural and regulatory components. As observed in nature, the developmental plasticity is a necessary condition for evolutionary processes -- such processes, which can potentially make a system more complex, increase information capacity and processing power~\cite{Ruiz-Mirazo08}.

Exploration of these issues represent a challenge for researchers and engineers. It is firstly related to a good engineering of mechatronic cell-modules, which should demonstrate 2D locomotion on a surface, 3D actuation within a heavy organism, autonomous docking to each other, large on-board energy resources, different sensors and sufficient computation/communication. Of utmost importance is that the modules should be small in size and light in weight. Not only mechatronics, but also software engendering and design of control and regulative structures are of essential importance. This paper is basically devoted to these challenges and represent a snapshot of the research and technological development conducted within the European projects ``SYMBRION"~\cite{symbrion} and ``REPLICATOR"~\cite{replicator}.

The paper is organized in the following way. The Sec.~\ref{sec:mechatronics} introduces development of heterogeneous reconfigurable platforms. Sec.~\ref{sec:electronicArchtecture} treats issues of general architecture, computational power and  on-board sensors, whereas Sec.~\ref{sec41:generalFramework} briefly considers the software framework. Finally, Sec.~\ref{sec:GrandChallenges} introduces one of the Grand Challenges and Sec.~\ref{sec:conclusion} concludes this work.

\section{Mechatronic Platforms}
\label{sec:mechatronics}

The mechanical characteristics and functionalities of individual robots in a collective symbiotic system are of the utmost importance in order to confer suitable capabilities to the symbiotic robot organisms. However, this does not necessarily mean that the design of individual robots has to be particularly complex from a mechanical point of view. On the contrary, excessive complexity can lead to several disadvantages in the assembled state of the organism, e.g. higher risk of failures and higher electrical and computational power demand. In addition, considering the manufacturing phase of the individual robots themselves, complexity would lead to high development and assembling costs; this is an issue particularly relevant when a large multi-agent symbiotic system is targeted. Finally, considering miniaturized robots, there are severe volume constraints at the design level that may prevent the possibility to integrate complex mechanisms. Consequently, as a rule of thumb, the individual robots of a large collective symbiotic system can be designed to offer the minimal mechanical functionalities able to allow the symbiotic robotic organism to assemble and develop all those \textit{collective configurations} and \textit{reconfiguration strategies} that let specific \textit{collective functionalities} emerge. That's inevitably a compromise choice in the design.

As already mentioned, a symbiotic robot organism can be seen as the physical evolution of a swarm system of individual robots into a structural system of connected robots. From this ``structural" perspective, the mechanical functionalities of the individual robot could correspond to the behavioral rules of the agents in a swarm system that generates collective emergent behaviors. The mechanical interactions between the robots assembled in the organism expand consequently the collective capabilities of the system to a structural dimension.

On the base of the above considerations, it is clear how the design of suitable mechanical features of the individual robots represents a critical issue. In particular, the robot-to-robot connection mechanisms (docking mechanisms) and the mechanical degrees of freedom implemented in the individual robots deserve a deep investigation.

\subsection{A Heterogeneous Approach in Modular Robotics}
\label{subsec:heterogenous_approach}

The design of each individual robot as a stand-alone unit inevitably ends to favor specific functional characteristics such as locomotion capability, actuation power and robustness, and this can result in multiple design solutions. This is true especially for miniaturized individual robots because focusing on one feature means finally to degrade or loose other features due to obvious space constraints.
As a consequence of the above mentioned issues, the design process can follow different paths:
\begin{itemize}
\item To try to merge the best features of all the conceived designs into a unique individual robot design by accepting performance compromises of the collective system while making the control of the organism easier. We refer to such a system as collective \textit{homogeneous} system. This is the path mostly followed by state-of-the art modular and reconfigurable robotics (~\cite{Yim2003},~\cite{Castano},~\cite{Christensen2006},~\cite{Zykov2007},~\cite{Kamimura05},~\cite{salemi2006sdm}, etc.).

\item To consider having two or more different individual robot types where each robot is optimised for specific functions. Each robot can assemble into a symbiotic organism by means of compatible docking units, thus empowering the global capabilities of the collective system in detriment of more complex control of the symbiotic organism due to its heterogeneity. We refer to such a system as a collective \textit{heterogeneous} system as introduced in~\cite{Lyder2008}.

\item To integrate ``tool modules" with the above mentioned collective \textit{homogeneous} system. Tool modules can be generally defined as devices whose functions are dedicated to a specific task. The tool modules can simply dock with the assembled organism, receive commands from the organism and possibly send data to the organism. These tool-modules could be, for instance, wheels, sensors, grippers, etc. By following this path, the system has to accept poor integration of the robot in favor of versatility. This approach is considered to be the evolved version of the collective \textit{homogeneous} system as demonstrated in~\cite{Zykov2008}.

\item To integrate ``tool modules" with the above mentioned collective \textit{heterogeneous} system. The main structure of the organism is composed of two or more different individual robots and the organism can be equipped with ``tool modules". The heterogeneity of the system becomes high, making the control more complex. The system is the most versatile and robust to the environment and given tasks. This is a rather new approach in modular robotics as studied in~\cite{Bordignon2008},~\cite{Kernbach08b},~\cite{replicator}.
\end{itemize}

Taking inspiration from the biological domain, it could be observed that natural swarms are often heterogeneous not only for the different behavioral specialization of each swarm member but also from a strict physical viewpoint (e.g., in a same colony there are insects with different physical capabilities, e.g. in ant colonies). However, differently from natural insect swarms, the conceived collective system should also be able to reach a collective structural level. This goal can be more complicated with heterogeneous individual robots, regarding the assembly process itself and, even more, for what concerns the onboard software (e.g., the self-learning and behavioral control of the symbiotic organism). As a case study, two individual robots, namely a Scout robot and Backbone robot, and one tool module, namely Active wheel, will be described hereafter and shown later in the chapter:
\begin{itemize}
\item A ``scout" robot equipped with far-range sensors and above all specialized in fast and flexible locomotion that can be used for inspection of the environment and for swift gathering of robots for the assembly. For this purpose, wheeled/caterpillar-like locomotion is advantageous, in particular where challenging terrains have to be engaged. Actuators for the 3D actuation within the organism is mandatory but less powerful actuators are sufficient. It is because the scout robots can be useful when they are docked to the end of a leg or arm of the organism to scan the environment.

\item A ``backbone" robot, strong in main actuation and stiff in design. The main purpose of this robot is to work as a part of the organism, therefore the casing is strong to provide high stability and the main actuator is able to lift several docked robots to perform 3D motion. The space for 2D locomotion is limited due to the large main actuator, but the 2D locomotion drive is capable of necessary movements for assembly and docking. In addition, the design of the robot allows to use the single DOF of the main actuator for either bending or rotation of the docked joint. Therefore, the powerful actuation is available for any joint in the assembled organism.

\item An ``active wheel" module as a tool module. Tool modules are optimised for specific functions and designed in a way to compensate aforementioned deficits of the individual robots. The Active wheel, for example, provides the ability to move omnidirectional, lifting and carrying heavy loads (i.e. other robots or organisms) and at the same time is able to provide an additional energy source. This tool can act in standalone mode as well as in organism mode.
\end{itemize}

\begin{figure}[htp]
\centering
\subfigure{\includegraphics[width=.49\textwidth]{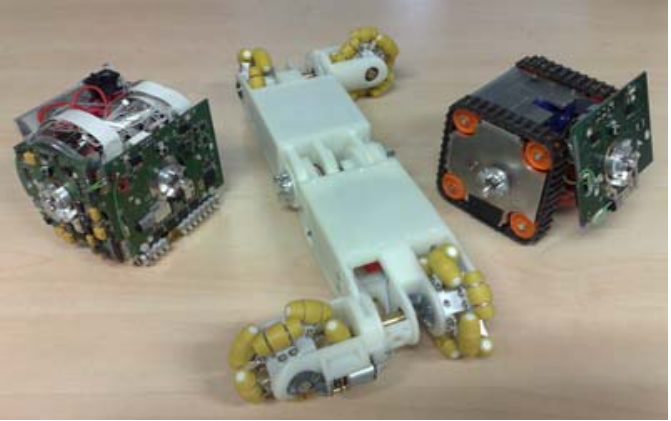}}
\caption{First prototypes of robot designs (from left to right): Backbone robot, Active wheel, and Scout robot.
\label{fig:sec31-Designs}}
\end{figure}

The prototypes of the Backbone robot, the Active wheel and the Scout robot are shown in Fig.~\ref{fig:sec31-Designs}.

Following the general issues introduced above, several technical key aspects have to be taken in consideration in the mechanical design of the individual robots.  The requirements and solutions of the Scout robot, the Backbone robot and the Active Wheel have been defined as shown in Table~\ref{tab:sec31-TwoRobots}.
\begin{table}[htbp]
\centering
\caption{Scout robot, Backbone robot and Active Wheel: requirements and solutions}
\label{tab:sec31-TwoRobots}
\begin{tabular}{p{.7cm}@{\extracolsep{3mm}}p{.8cm}@{\extracolsep{4mm}}p{.8cm}@{\extracolsep{3mm}}p{1.cm}@{\extracolsep{3mm}}p{1.3cm}@{\extracolsep{3mm}}p{.8cm}@{\extracolsep{3mm}}p{.8cm}}
\hline\noalign{\smallskip}
& \multicolumn{2}{c}{Scout robot} & \multicolumn{2}{c}{Backbone robot} & \multicolumn{2}{c}{Active Wheel}\\
\hline
									& Require.      & Solut. 	       & Require.     & Solut.   & Require.   & Solut.
 \\\noalign{\smallskip}\hline\noalign{\smallskip}

Align-ment 				  & Rough & Tracked locomotion & Accurate & Omni- directional drive &Accurate &Omni- directional\\
Ground Surface 		& Rough & Tracked locomotion & Plain 	& nearly Omni- directional &Plain &Omni- directional\\
Locom.
after docking       & Required to carry a robot		 & OK (3 surfaces)	& Not required                                      & wheels still available for driving & Required to carry & OK (2 surfaces)\\
Speed, loc.  & High 		& 12.5 cm/s 				 & Low 			& 6 cm/s & High 		& 31 cm/s\\
%that is not completely true
DOFs of actuation  	& 2 DOF 	& Bending: $\pm90^\circ$ Rot.: $\pm180^\circ$ & 1 DOF 		                                                          & Bending/ Rot.: $\pm90^\circ$  & 2 DOF  & Bending/ Rot.: $\pm180^\circ$  \\
%Torque 						& Low 2.9 Nm: Lifting up 2 docked robots  & 3.2 Nm (Nominal)	& High 	& up to 6N  \\
Torque 							& Low 		& 3Nm								 & High 		& up to 7Nm & High 		& up to 5Nm \\
Speed, act. 	& Low $30^\circ$/s& $37.2^\circ$/s & High 			 	& $180^\circ$/s & Low 			 	 & $50^\circ$/s\\
\noalign{\smallskip}\hline\noalign{\smallskip}
\end{tabular}
\end{table}

\subsection{Locomotion Mechanisms of Backbone and Scout Robots}
\label{subsec:locomotion_mechanisms}

The locomotion capability allows the individual robots to be active in the environment, carrying on tasks of exploration, for instance. The locomotion capability is evidently fundamental when docking with other robots is necessary in order to reach the symbiotic state. Several approaches can be followed for the design of locomotion mechanisms, depending on the requirements that the individual robots and the symbiotic organism have. In classical modular robotics, the individual robot or module has been considered as a part of the modular system, thus it does not have any mechanisms that let it move as a stand-alone system. Instead, locomotion has generally been considered as a capability of the assembled robot and achieved by means of coordinated actuation among the docked modules in order to realize snake-like locomotion, legged-base walking, etc. This can limit the exploration capability of the whole system to the assembled state. In other words, individual robots or modules need to be manually positioned and docked before initiating the operation. When additional modules are requested by an assembled robot at the operation site, the assembled robot needs to go back to a specific zone where individual modules are deployed, or another assembled robot needs to be formed to reach the operation site. Hence, it is a natural consequence to try to devise individual locomotion solutions on each individual robot. This would guarantee the collective system much higher independence, versatility and flexibility. The system can be autonomous and robust especially in an unknown environment where the number of required robots and appropriate topologies of the organism can be determined after the robots reach the operation site.

\begin{figure}[htp]
\centering
\subfigure[]{\includegraphics[width=.24\textwidth]{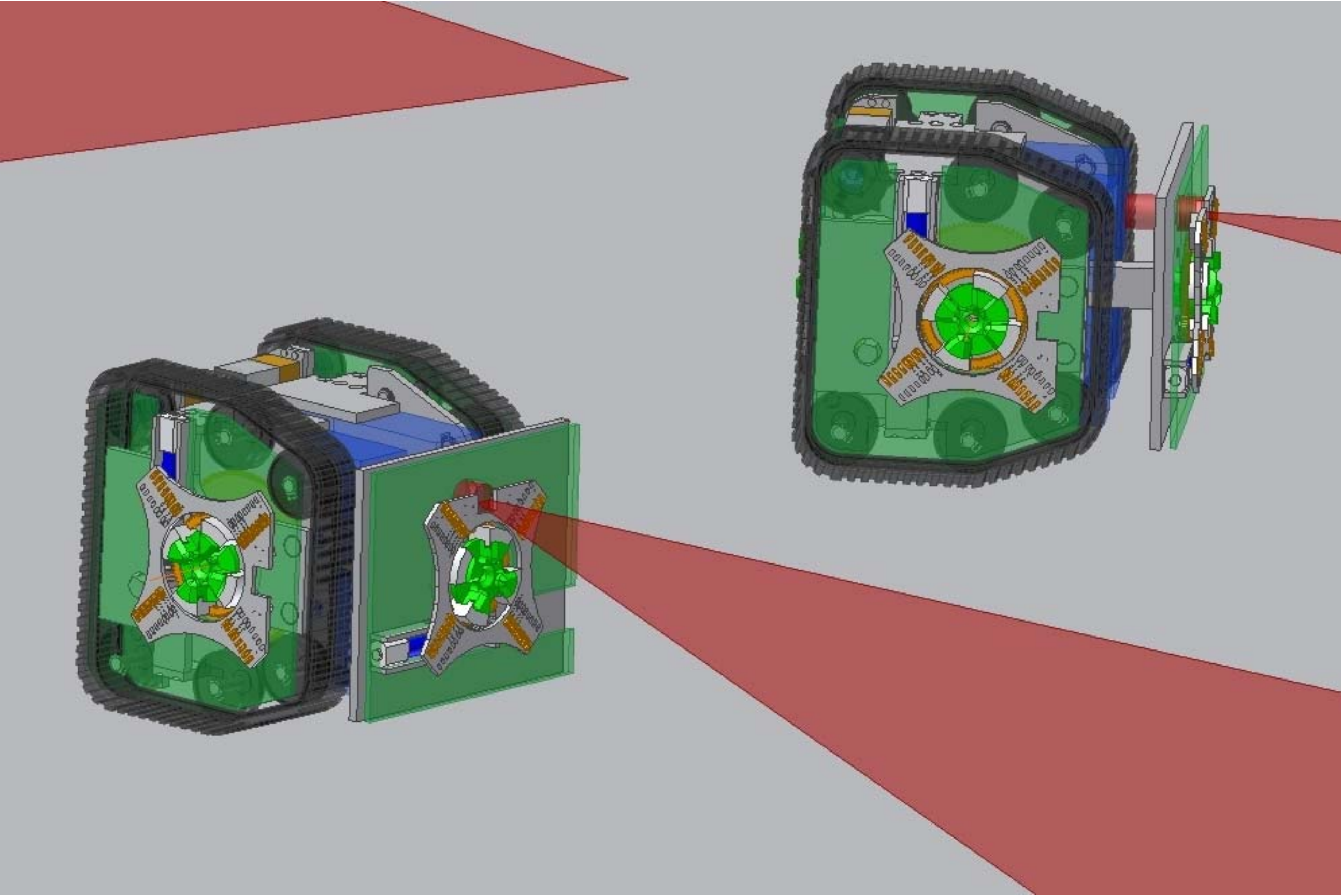}}~
\subfigure[]{\includegraphics[width=.24\textwidth]{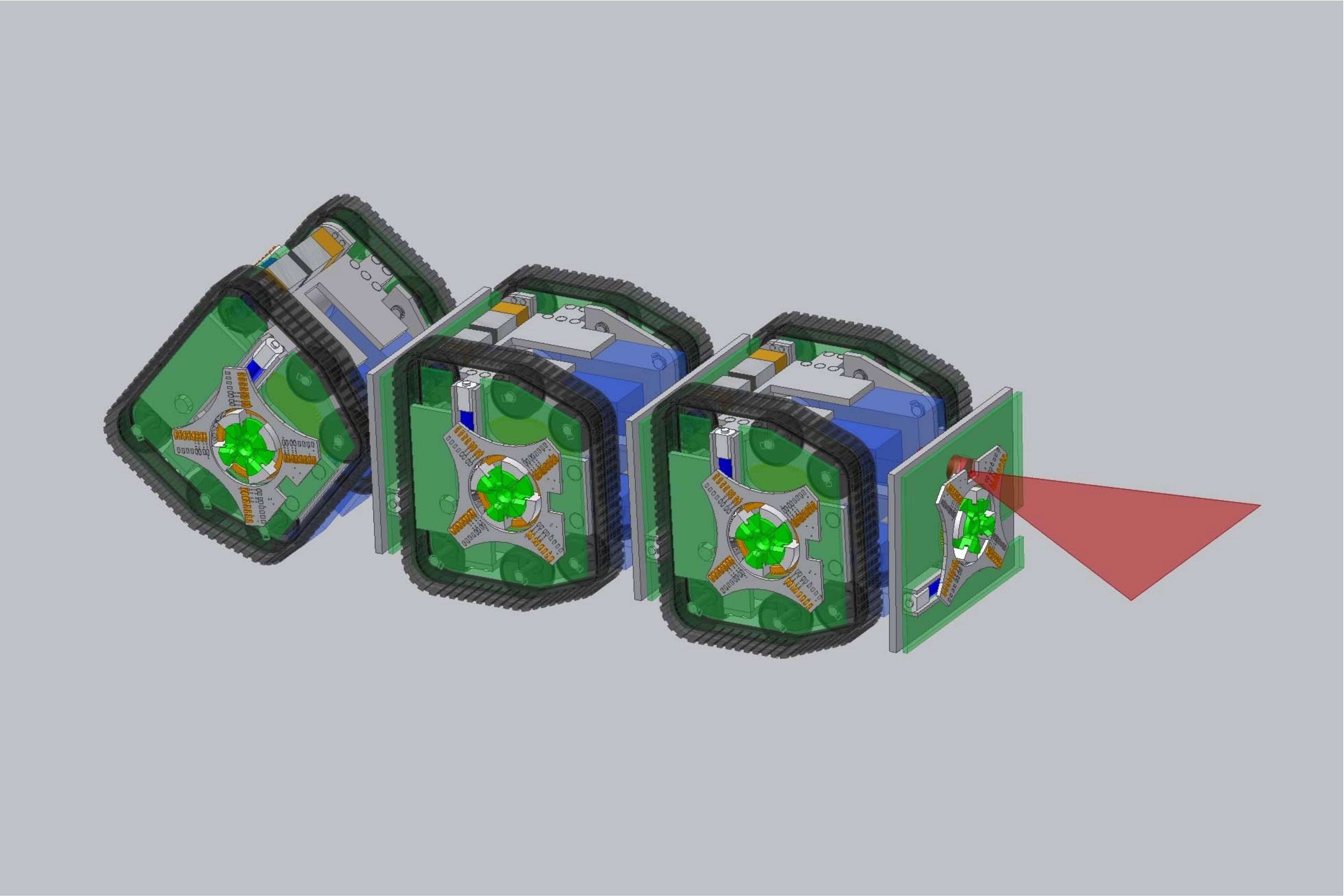}}\\
\subfigure[]{\includegraphics[width=.24\textwidth]{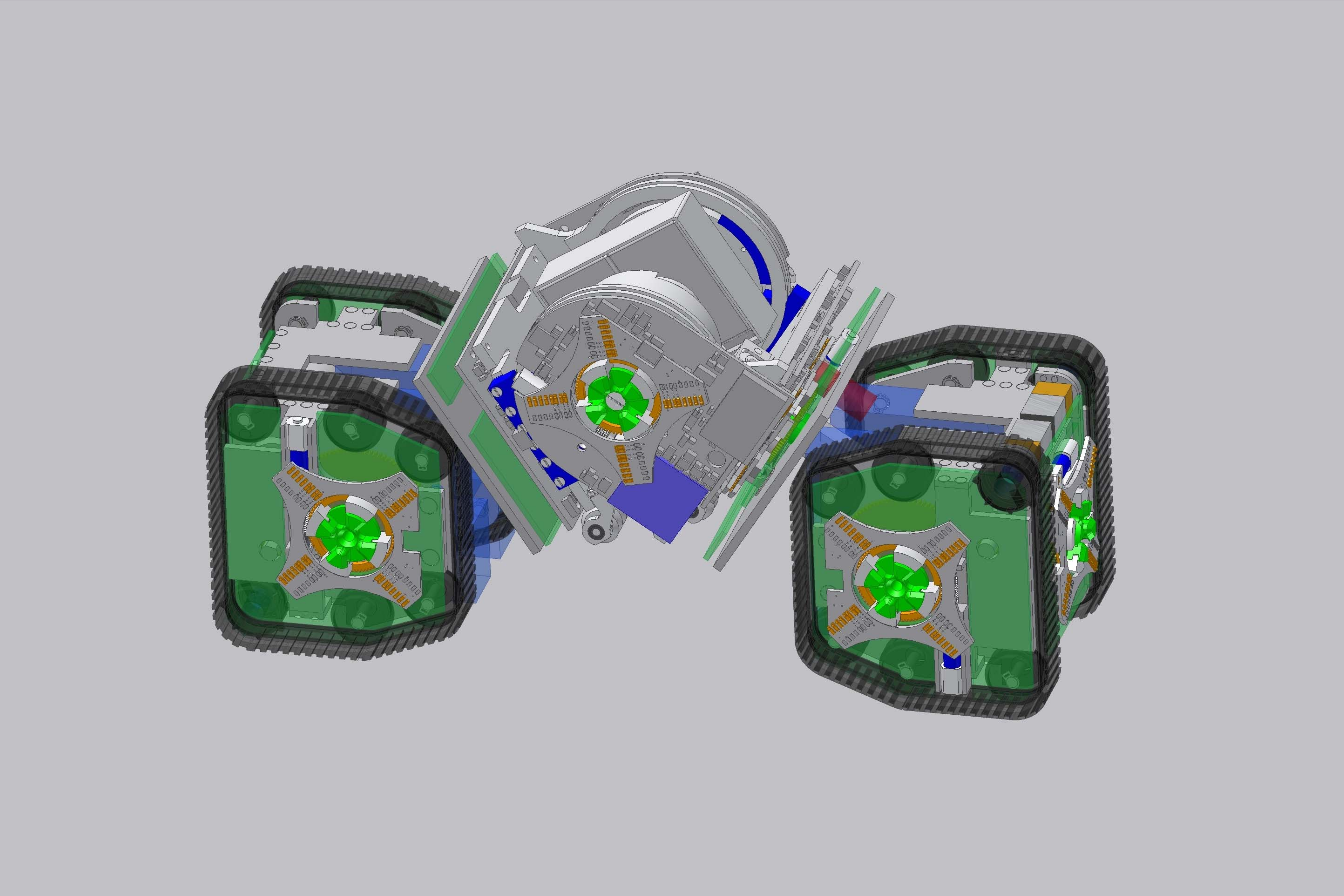}}~
\subfigure[]{\includegraphics[width=.24\textwidth]{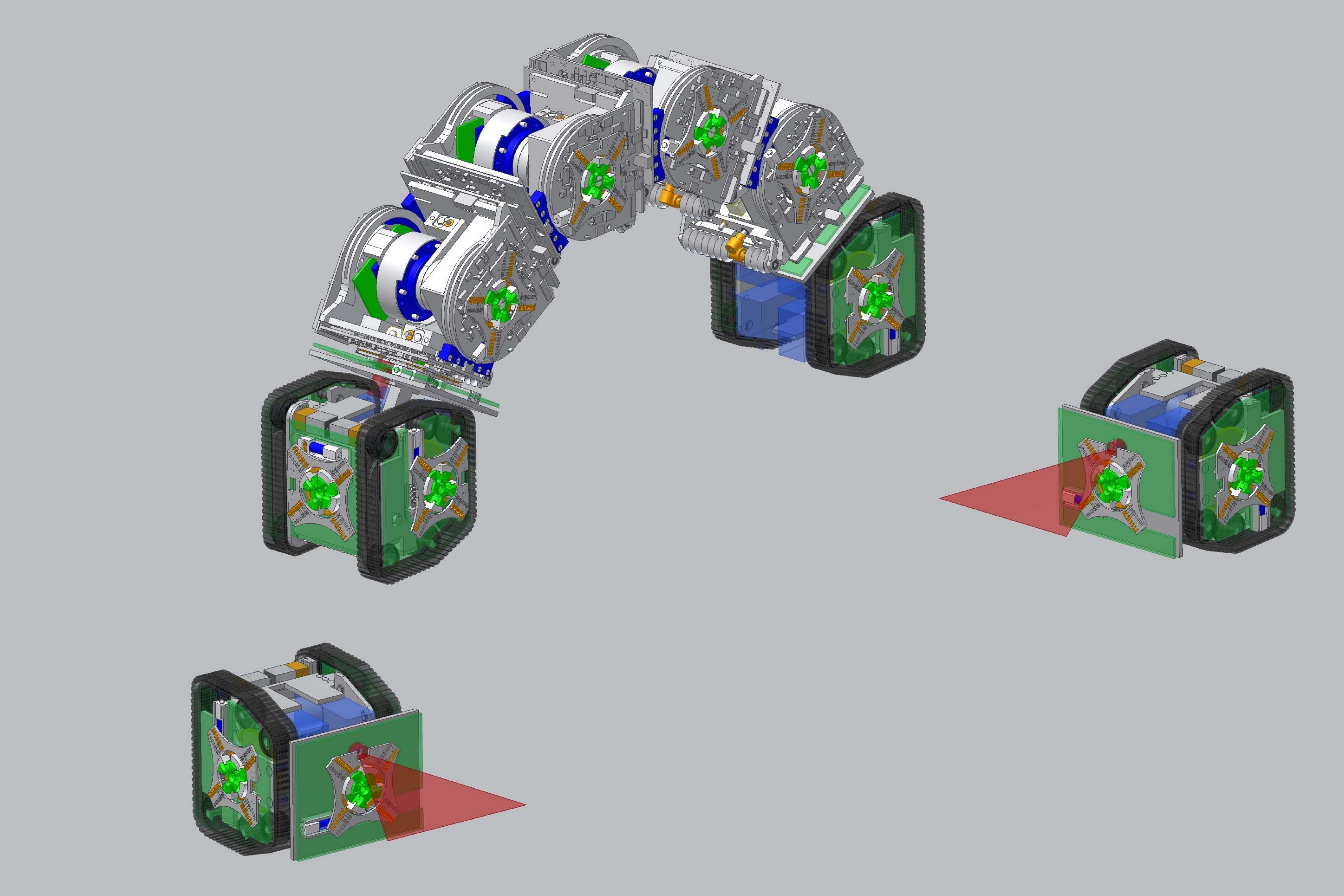}}\\
\subfigure[]{\includegraphics[width=.24\textwidth]{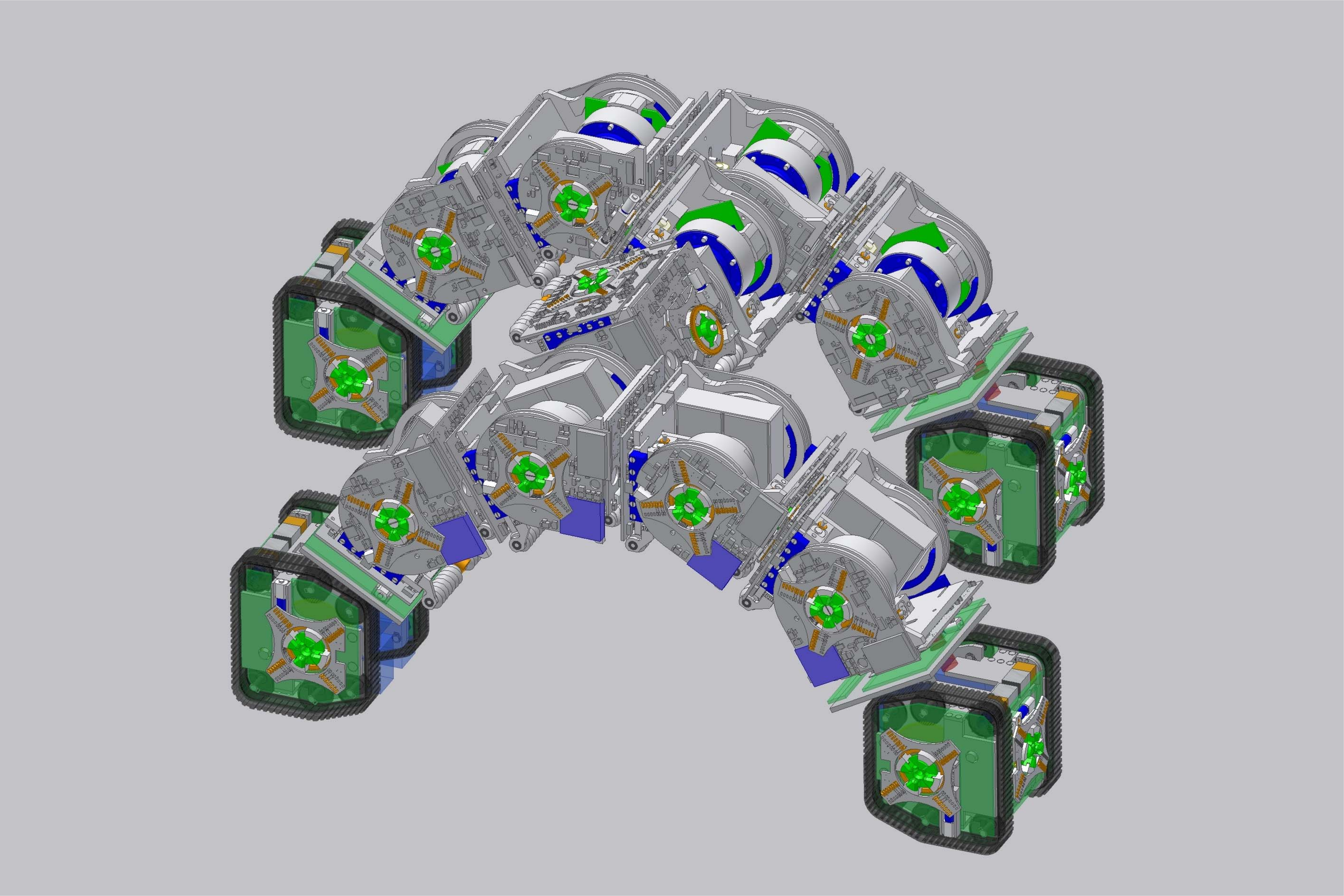}}~
\subfigure[]{\includegraphics[width=.24\textwidth]{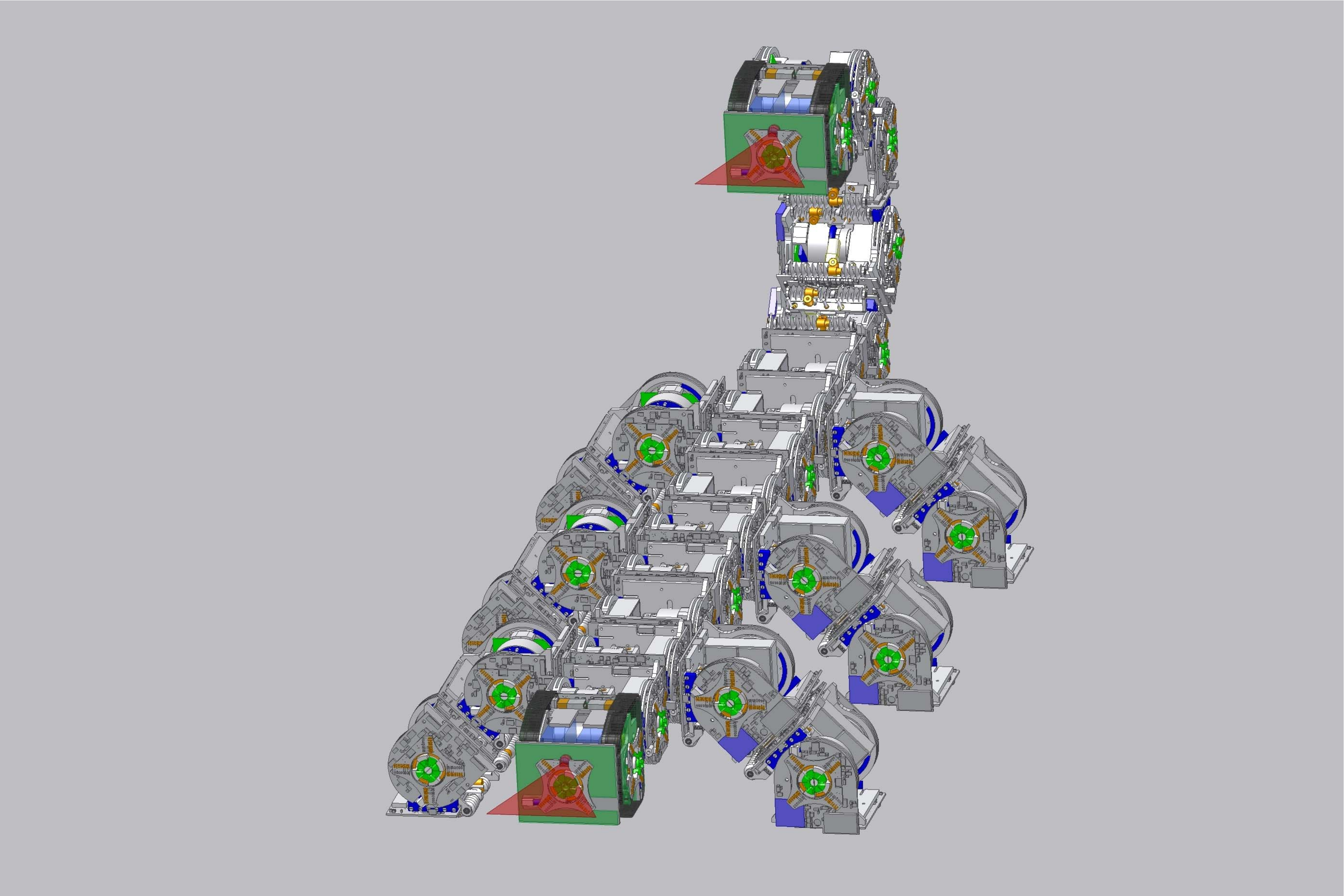}}
\caption{Scout robots:
\textbf{(a)} Scout robots exploring the surface and guiding the organisms;
\textbf{(b)} Connected Scout robot;
\textbf{(c)} Scout robots carrying a Backbone robot;
\textbf{(d)} Scout robots carrying a chain composed of the Backbone robots.
\textbf{(e)} 4-legs shape of an organism;
\textbf{(f)} Scorpion-like organism.
\label{fig:sec31-ScoutRobot}
\label{fig:sec31-ScoutRobotCarry}
\label{fig:sec31-Organisms}}
\end{figure}

Tracked locomotion is adequate for the quick locomotion on rough terrains. The Scout robot with tracked locomotion is capable of going up a slight slope, climbing over small obstacles, passing over a small hole, and also moving in soft ground. The long-range sensors on board can be used to scan the obstacles around then to navigate the organisms (Fig.~\ref{fig:sec31-ScoutRobot}(a)). When the tracked robots are docked together, the assembled robot becomes more robust to the roughness of the terrains as shown in Fig.~\ref{fig:sec31-ScoutRobot}(b). This high locomotive capability also allows the Scout robots to carry the Backbone robot(s) (see Figs.~\ref{fig:sec31-ScoutRobotCarry}(c)(d)). The Backbone robots can form an arm or a leg of an organism in advance, then be carried to the operation site so as to save the energy for 3D actuation in the organism. Thus, the Scout robots are adequate to be ``feet" of the organism thanks to their robustness and locomotive capability. The disadvantage of the tracked locomotion is the non-holonomic drive characteristic that hinders efficient docking procedures between the robots.

Regarding the locomotion capability of the Backbone robot, easy assembly of the organism is of utmost importance. Therefore an omnidirectional drive is best since it offers optimal performance to move to a predefined position under a defined angle. This is important because each individual robot provides at least four different docking units and all of them can be used to form the structure of the organism. Every docking unit needs to be reached, regardless of the orientation of the robot which wants to dock. Unfortunately, the integration of an omnidirectional drive requires a lot of space due to the general construction of omnidirectional wheels. Nevertheless, if one takes a closer look at the details of the docking procedure, complete omnidirectional driving characteristics are not required for the Backbone robot, since the orientation of the robot is predefined by the docking units and therefore only certain directions of movement are necessary. In general, the Backbone robot needs to be able to move forward, backward and to turn since these are the minimum requirements for a swarm robot. Furthermore, under the condition of docking orthogonally to the normal drive direction of the robot, it needs to move sideways. A locomotion drive unit which can provide the features of a differential drive plus the possibility to drive to the side is therefore sufficient. Both features are provided by the screw drive, which is used within the Backbone robot. The screw drive locomotion unit itself can be built very small since only two driving motors are required and the driving screws have cylindrical shapes.

Beyond the normal use of the nearly omnidirectional drive of the Backbone robot, the screw drive provides the organism with a possibility to move sideways when the screws of all robots within the organism are synchronised. This can be a very helpful feature if a caterpillar like organism needs to steer to the side. An example of a system composed of reconfigurable heterogeneous mechanical modules, i.e. the Scout robots and the Backbone robots, are shown in the Figs.~\ref{fig:sec31-Organisms}(e)-(g). All individual robots and organisms work as autonomous stand-alone systems.

\subsection{Tool module: Active Wheel}
\label{subsec:tools}

In a heterogeneous system, robots of different design can form an organism together. The two individual robots, namely Scout robot and Backbone robot, have been proposed as basic elements to constitute an organism. The design of this individual robot is a result of compromise to integrate all mechanical and electronic functions into one robot. The features of such individual robots have to be redundant to be adaptable in an unknown environment. The idea of implementing tool modules into the heterogeneous system is to provide a few specially designed tools to compensate for deficits of the individual robots. The design of tool modules needs to be optimized for specific tasks such as sensing with a special sensor, manipulating an object, supplying power to the organism and carrying the individual robots or an organism quickly. The individual robots need to share external dimensions to be a part of the organism and for easy reconfiguration, and they need to be equipped with common electronics, while a tool module may have any shape as long as it can be docked to other individual robots or an organism.
As an example of tool modules, we developed a tool module to carry individual robots, named Active Wheel (see Fig.~\ref{fig:sec31-Designs}). This tool module is intended to carry some individual robots quickly from one place to another without using their energy. The Active Wheel is an autonomous tool robot that is compatible with the other two individual robots platforms (Scout robot and Backbone robot) and used for assistance goals. An Active Wheel consists of two symmetrical arms connected in the middle by a hinge.

\begin{figure}[htp]
\centering
\subfigure{\includegraphics[width=.48\textwidth]{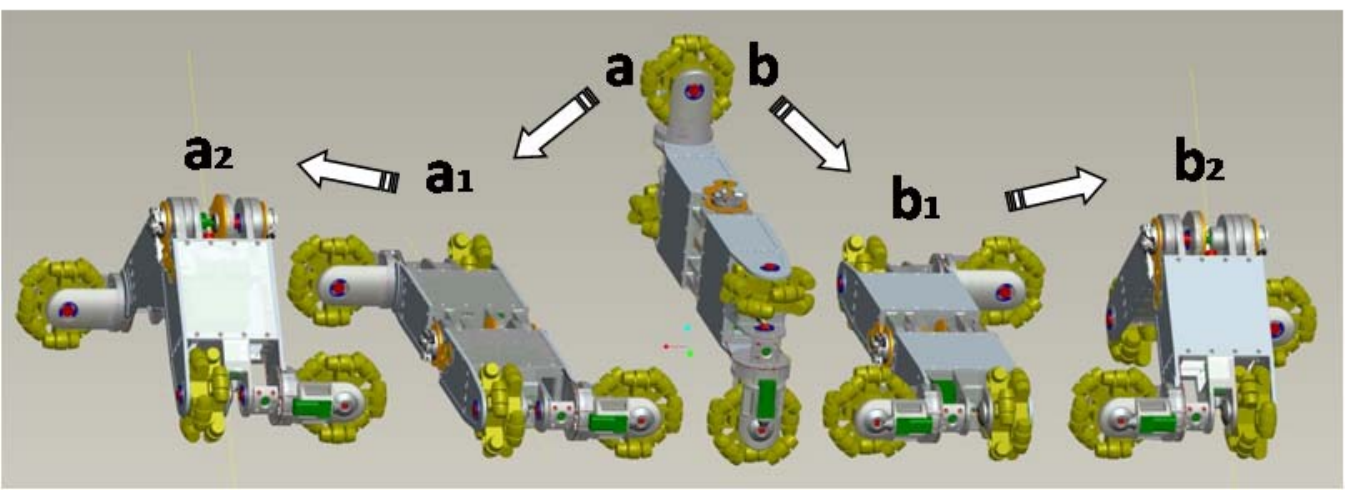}}
\caption{Symmetry and stability of the robot and capability to bend upwards or downwards.
\label{fig:sec31-Tools2}}
\end{figure}

This structure gives the opportunity of bending this tool in both directions up to $\pm$90$^\circ$ and hence can drive even upside down. Actually, such a symmetrical design does not require distinguishing between bottom and top or between front and rear side. An additional advantage of this geometry is the uniform weight distribution which is important for stable locomotion. Even if the robot is in a skew position $a$ or $b$ it tilts autonomously back into a stable position $a_1$ or $b_1$ (Fig.~\ref{fig:sec31-Tools2}).
One of the major tasks of this tool robot is to carry a certain number of individual robots efficiently from one place to another.
This condition can be fulfilled only if the Active Wheel can move omnidirectionally. Therefore, two omnidirectional wheels are used on each side on the robot. Such kind of wheels have already been  proven to work reliably in many robotics projects e.g. in RoboCup ~\cite{Zweigle09}. Each wheel consists of many small single rolls which are arranged perpendicularly to the driving axle.
This assembly allows an active movement in the driving direction of the wheel and simultaneously allows a passive movement in the normal direction. Each of these wheels is driven by a gear motor. Corresponding sensors which are placed on the driving axle detect the rotation speed of the motor. Those are necessary in order to provide complex manoeuvres such as driving curves or other complex trajectories. The docking between Active Wheel and another robot requires also a very precise control of the wheels.

Additionally to the motor control unit, the Active Wheel is equipped with similar electronic units and components like in the Scout or in the Backbone robot. These comprise for example similar processors, power management, IR sensing units, a ZigBee module, cameras etc. All these electronics are mainly required in order to navigate and to transport other robots autonomously and at the same time allow acting as stand-alone robot and fulfill many different tasks in robot swarms.
In stand-alone mode, Active Wheels can be used for separating damaged modules or modules that are not able to move.
\begin{figure}[htp]
\centering
\subfigure{\includegraphics[width=.48\textwidth]{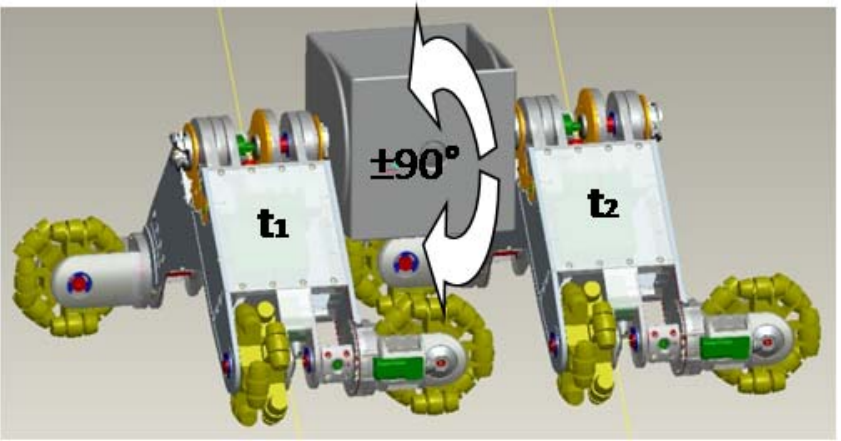}}
\caption{Two Active Wheels carry a defective element.
\label{fig:sec31-Tools3}}
\end{figure}
One possible scenario how an Active Wheel can act as a stand-alone robot, is shown in Fig.~\ref{fig:sec31-Tools3}. Two  Active Wheels are placing a module that was flipped over in the right position again.

\begin{figure}[htp]
\centering
\subfigure{\includegraphics[width=.48\textwidth]{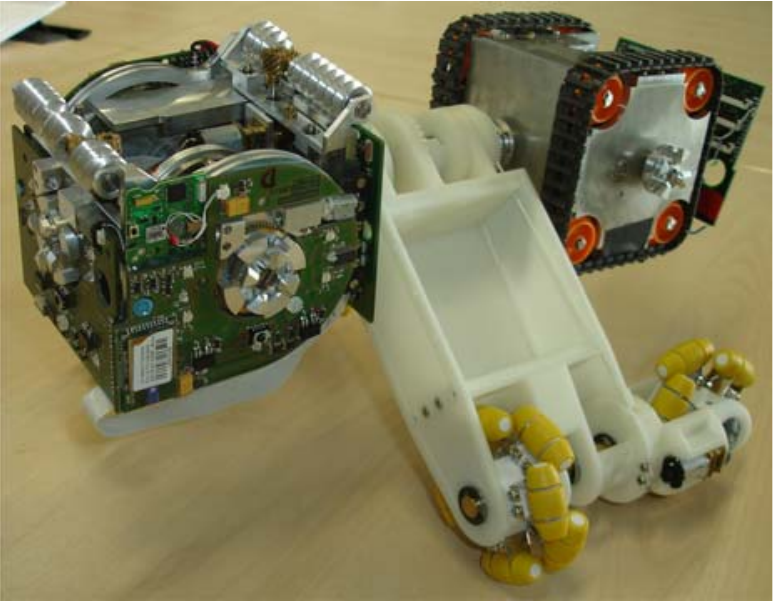}}
\caption{Simple organism - Active Wheels with two different docked modules.
\label{fig:sec31-Tools4}}
\end{figure}

As an example of a simple organism, topology of three robots can be considered Fig.~\ref{fig:sec31-Tools4}. The idea of this configuration is based on a combination of advanced computational and sensor features, provided by these two individual robots, and fast motion speed, provided by the Active Wheel.
Additionally, the Active Wheel can supply both individual robots with extended energy source. As a common system, these three platforms complement each other and demonstrate commonly very outstanding characteristics. Features of a common system essentially excel the capability of each of these individual robots -- this is typically the collective approach.

\subsection{Docking Mechanisms and Strategies}
\label{subsec:docking_mechanisms_strategies}

The docking mechanisms are of primary importance in modular robotics as well as in symbiotic multi-robot organisms. They should assure docking and undocking between individual robots, as well as electrical continuity for power sharing and signal transmission. Furthermore, the docking mechanism should tolerate at a certain degree misalignments of individual robots during the docking process~\cite{salemi2006sdm}. Nilsson et al. have investigated design of a docking unit and summarized desirable connector properties~\cite{Nilsson2002}. In this section, the properties required for docking mechanisms are investigated and a guideline for the docking design is proposed. \textit{Docking} is composed of several phases, and each phase has several requirements to be satisfied.

\textbf{Approach}. The approach of the docking units can be categorized into three modes. The first is the approach of the two locomotive individual robots. Because both robots can move freely, the approach of the docking units is rather easy. The second is the approach of an individual robot to an organism. In this case, the individual robot should be precisely steered. When the individual robot with non-holonomic locomotion capability needs to be docked to the organism, the docking units on the side walls are not available unless the organism itself can approach the individual robot. Thus, the aggregation of an organism must be carefully planned considering the locomotion capability of the individual robots. The last one is the approach of the two assembled robots or two arms/legs of an organism, and this is especially important for a reconfiguration of the organism.

\textbf{Alignment}. Docking design that allows robust self-alignment is crucial for autonomous assembly of a modular robot. Ground roughness needs to be taken into consideration for the docking of locomotive individual robots. In addition, it must be noted that the accuracy of the fabrication and assembly of each robot hase strong influence on the alignment accuracy.

\textbf{Docking and Locking}. A docking unit with hermaphroditic feature is preferable to make the assembly plan easier. The docking must be tight and stable, and the electrical connection between the docked robots must be ensured. In some existing docking designs, the docking is secured by an additional locking mechanism. A simple docking/locking mechanism occupying small space and being actuated with little energy is preferable as well.

\textbf{Sustainment of the docked status}. The docking status must be sustained without or with minimum power supply. The docking status needs to be independent from the actuation of the assembled robots, otherwise, the additional control is necessary to maintain the docking status.

\textbf{Unlocking and Undocking}. Another important feature is the capability to allow undocking between two docked robots in case of an emergency. If one of the individual robots undergoes failure or malfunction, the robot must be removed from the organism by the other robots. Therefore, it is preferable to undock the robot by activating only one of the docked units.

\textbf{Separation}. The individual robots need to be separated and move away from the assembled robot after being undocked so as not to hinder following procedures. When an individual robot with non-holonomic locomotion cannot move away after being undocked, the organism needs to move away from it or another robot needs to come to move it away.

In addition to the above mentioned requirements, easy and low-cost manufacturing for mass production and easy maintenance are important especially when a large multi-agent symbiotic system is targeted. Because multiple docking units are required for an individual robot, the cost of the docking unit is important.

To summarize this section, we have to point out two essential issues: integration with electronics, and a need of software protection from mechanical damages, caused during evolving different controllers. Both issues are essential in a successful design and stepwise improvement of mechatronic platforms.

\section{General Hardware Architecture}
\label{sec:electronicArchtecture}

\begin{figure}
	\centering
		 \includegraphics[width=0.5\textwidth]{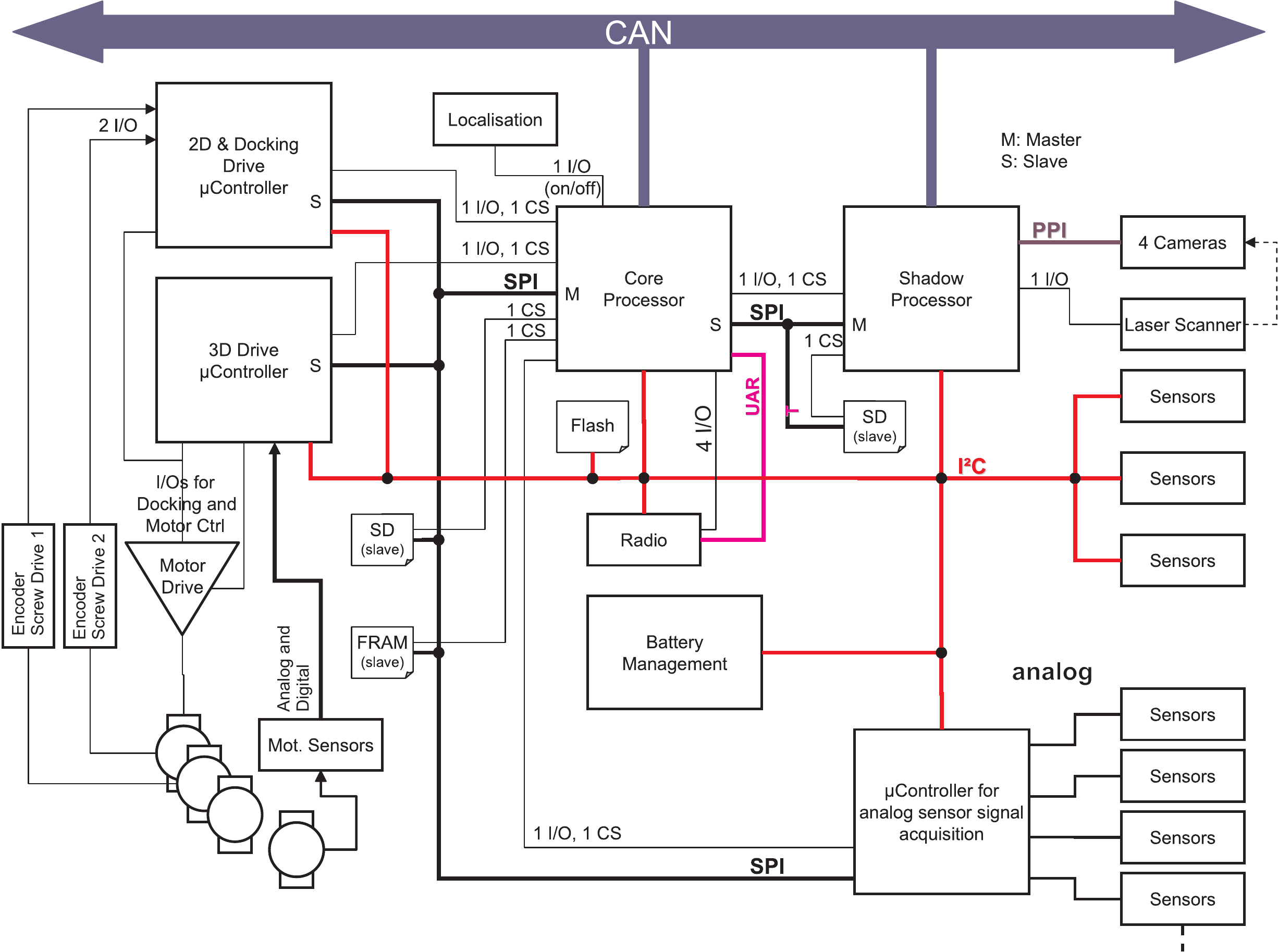}
	\caption{Electronic architecture of the \rep\ robotic modules.}
	\label{fig:BlockDiagramArchitecture}
\end{figure}

In this section, the electronic hardware and architecture of single robot modules (the first prototype) is described in more detail as another example of self-reconfigurable robots (see Fig. \ref{fig:BlockDiagramArchitecture}). Since in \textsc{Symbrion} advanced control and evolutionary algorithms, such as on-board genetic evolving, etc. needed to be implemented, here, one major design criterion was the calculation and processing speed. On the other hand, \textsc{Replicator} required a high number of different sensors since the swarm's objective was to form a highly dynamic sensor network for vast applications, like surveillance, exploration, etc.
As shown in Fig.~\ref{fig:BlockDiagramArchitecture}, each module hence carries a number of processors/microcontrollers. However the major control of each robot is performed by the ``Core Processor'', an \textit{LM3S8970} \textit{Cortex} microcontroller from \textsc{Luminary Micro Inc}. The main purpose of it is to pre-process raw sensor data, to run higher level algorithms such as an artificial immune system (AIS) or artificial homeostatic hormone system (AHHS), to calculate the module's position, to pass this information to actuators, etc. In order to support this processor, a shadow processor (\textit{Blackfin}, \textit{ADSP-BF537E} from \textsc{Analog Devices}) is included that mainly takes over computationally intensive processing tasks, i.e. of the images taken from the 4 on-board cameras. Due to its high power consumption, the intention is to operate this processor unit only if required. For example, if image processing has to be used to recognize the environment or if the organism size (i.e. number of docked modules) reaches a certain limit so that locomotion tasks require a lot more computational resources.

A dedicated microcontroller (\textit{ATmega1280} from \textsc{Atmel Inc.}) is responsible for A/D-conversion and further processing of analogue sensor signals like microphones, IR-based distance sensors, etc. Since at least 1 brushless motor, whose control occupies many processing resources, is on board a robot module 2 additional Cortex controllers (\textit{LM3S8962}) have been integrated, dedicated to all major actuation and locomotion tasks. Furthermore, the robots possess a UWB-based localisation unit, a \textit{ZigBee}$^{\mbox{\tiny TM}}$ radio communication module, a battery management module, Flash and SD memory, a LASER ranging module, and other sensors.

\subsection{General Sensor Capabilities}
\label{sec:electronicsFitness}

Following the approach from the previous section, we consider now the general sensor capabilities of the platform. For the application of evolutionary approaches as well as for sensor network applications, the platform should provide a measurement of environmental values, in particular, how robots do fit to the environment. The local fitness measurement for collective behavior represents a very challenging task, therefore a serious attention during the design of the platform was paid to this issue. From a conceptual viewpoint, the following four ways are available to measure the fitness: approximation of a global state by local sensors,  perception of local environment by on-board sensors, different measurements during robot-robot interaction, and finally, measurements of internal states.
\begin{table}[htb]
\caption {Overview of on-board sensors. \label{tab:sensorsImplemented}}
\centering
\begin{tabular}{p{3.7cm}@{\extracolsep{3mm}}p{2cm}@{\extracolsep{3mm}}p{1.5cm}}
\hline\noalign{\smallskip}
 Sensor & Name & Interface \\\noalign{\smallskip}\hline\noalign{\smallskip}
 \textbf{Environmental}\\
 Light & ADPS9002 & analog \\
 Air Pressure & SCP1000 & I2C \\
 Directional Sound & SPM0208HD5 & analog  \\
 Humidity/Temper. & SHT15 & I2C  \\
 IR-reflective & TCRT1000 & analog  \\
 Imaging Sensor & OV7660FSL & PPI \\
 Laser (in the Range Finder)& LS-1-650 & digital\\
 RFID sensor & Lux & no \\
 Sonar sensor & SRF08(or 10) & I2C  \\
 Laser RangeFinder & URG-04LX & RS232/USB  \\
 Detecting motion & AMN34111 & analog \\
 Hall effect (magnetic) & US4881EUA & analog \\
 Color Sensor & TCS230 & digital \\
 Capacitive & MT0.1N-NR & digital \\
 \textbf{Locomotion}\\
 3D Acceleration & LIS3L02AL & I2C  \\
 WTL laser mouse & ADNS-7530 & SPI \\
 3D Localization & Ubisense & digital  \\
 Orientation-Sensor & SFH 7710 & SPI \\
 IR-docking sensor & IR-based & analog\\
 Force measurement sensor &K100N & analog \\
 Joint angle sensor & 2SA-10-LPCC & analog  \\
 Compass & HMC5843 & digital \\
 \textbf{Internal, Indirect Sensors} \\
 Voltage, Current & BQ77PL900DL & SMBus \\
 Bus Load Sensor & no & software \\
 Center of mass & no & software \\
 Energy-docking sen.& no & software \\
\noalign{\smallskip}\hline\noalign{\smallskip}
\end{tabular}
\end{table}

\textbf{1. Approximation of a global state by local sensors.} For an application of evolutionary strategies the most appropriate feedback may be provided when knowing a global state of the environment, including internal states of other robots. However, such information is not available for individual robots due to practical reasons. Nevertheless, the global state can be approximated when using the world model and several sensor-fusion approaches. Examples of global states are map-related values, such as explored/unexplored area, coverage of some territory, position of robots in 3D space. The platform includes several sensors, such as localization system or laser rangers, for these purposes.

\textbf{2. Sensing a local environment.} Perception of local environment by on-board sensors is the primary way of receiving information about the environment for both evolving and sensor network applications. The overview of integrated, or considered for integration, sensors is given in Table~\ref{tab:sensorsImplemented}.

\textbf{3. Information provided by a robot-robot interaction and communication.} Robot-robot interaction is a very important source of fitness measurement. The corresponding sensors are the force measurement sensors, joint angle, compass or 3D accelerations.   Robot-robot communication plays also an important role here, which allows fusing local information from different robots. This is related not only to environmental values, but also to internal states of robots.

\textbf{4. Internal states of robot organisms.} There are different internal sources of information: energy-based, mechanical, load on buses, number of internal failures, CPU/Memory usage and other. The energy-based values are very useful for many purposes, e.g. in estimation of the most efficient structure of organisms. Generally, the number of internal sensors, most of them are virtual sensors, can be very high.

To give a reader an impression about sensing capabilities of the platform, we collect in Table~\ref{tab:sensorsImplemented} a brief overview of on-board sensors.

\section{Controller Framework}
\label{sec41:generalFramework}

\begin{figure}[htp]
\centering
\includegraphics[width=.45\textwidth]{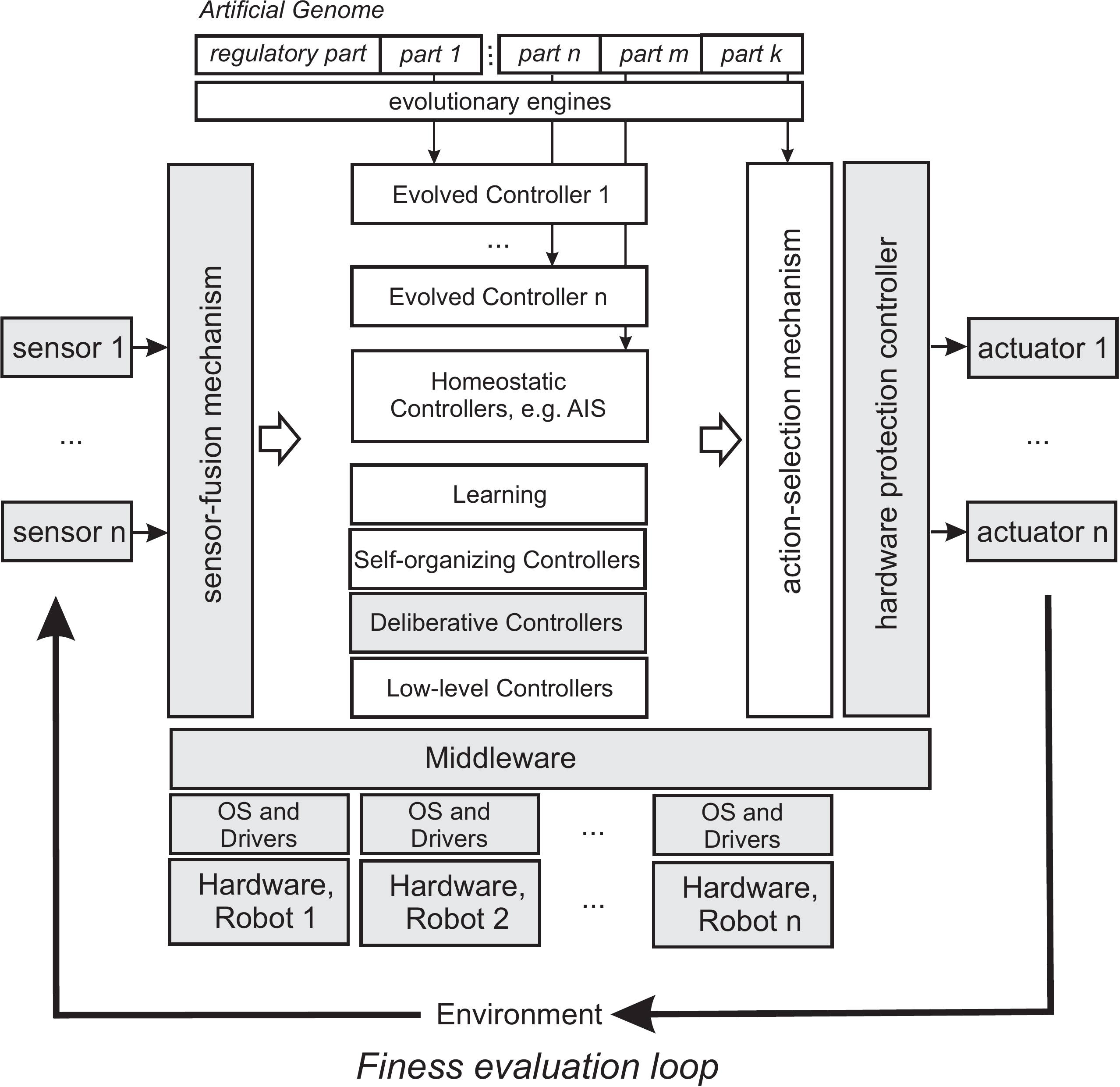}
\caption{General controller framework. All controllers/processes are distributed in the computational system of an artificial organism, OS -- operating system. Structure of controllers utilizes hybrid deliberative/reactive principle. \label{fig:sec41-generalFramework}}
\end{figure}

In robotics, several different control architectures are well-known, as e.g. subsumption/reactive architectures~\cite{Brooks-subsumption86}, insect-based schemes~\cite{Chiel92} or structural, synchronous/asynchronous schemes, e.g.~\cite{Reid91}. An overview of these and other architectures can be found in \cite{Siciliano08}. Recently, multiple bio-inspired and swarm-optimized control architectures have appeared, e.g.~\cite{Tianyun08},~\cite{Kornienko_S06}. In designing the general control architecture, we face several essential challenges:
\begin{itemize}
  \item \textbf{Multiple processes.} Artificial organisms execute many different processes, such as evolutionary development, homeostasis and self - organizing control, learning, middle- and low-level management of software and hardware structures. Several of these processes require simultaneous access to hardware or should be executed under real-time conditions.
  \item \textbf{Distributed execution.} Hardware provides several low-power and high-power microcontrollers and microprocessors in one robot module. Moreover, all modules communicate through a high-speed bus. Thus, the multiprocessor distributed system of an artificial organism provides essential computational resources, however their synchronization and management present a challenge.
  \item \textbf{Multiple fitness.} Fitness evaluation by using local sensors is already mentioned in Sect.~\ref{sec:electronicsFitness}. Here we need to mention the problem of credit assignment related to the identification of a responsible controller, see e.g.~\cite{Whitacre06}. Since many different controllers are simultaneously running on-board, the problem of credit assignment as well as \emph{interference between controllers} is vital.
  \item \textbf{Hardware protection.} Since several controllers use the trial-and-error principle, the hardware of robot platform should be protected from possible damage caused during the controllers' evolution.
\end{itemize}
Corresponding to the hardware architecture, the general controller framework is shown in Fig.~\ref{fig:sec41-generalFramework}.
This structure follows the design principles, originating from \emph{hybrid deliberative/reactive systems}, see e.g.~\cite{Arkin94planningto}. It includes a strongly rule-based control component, see e.g.~\cite{Li06} as well as multiple adaptive components~\cite{kernbach09adaptive}. The advantage of the hybrid architecture is that it combines evolvability of reactive controllers, and their high adaptive potential, with deliberative controllers that provide planning and reasoning approaches required for the complex activities of an artificial organism.

Controllers are started as independent computational processes, which can communicate with each other and with different sensor-fusion mechanisms, such as virtual sensors or the world model. Processes are running on different modules, synchronization and interaction between them is performed through message-based middleware system. There are controllers, which use evolutionary engines and their structure is coded in the artificial genome. There are several bio-inspired ideas towards such an artificial genome. It is assumed that there are also a few task-specific controllers, which are placed hierarchically higher than other controllers. These task-specific controllers are in charge of the macroscopic control of an artificial organism. They may use deliberative architectures with different planning approaches, see e.g.~\cite{Weiss99}.

The action-selection mechanism is one of the most complex elements of the general controller framework. This mechanism reflects a common problem of intelligent systems, i.e. ``what to do next'', see~\cite{Bratman87}. Finally, a hardware protection controller closes the fitness evaluation loop for the evolvable part of controllers~\cite{Kernbach09_CEC}. This controller has a reactive character and monitors activities between the action-selection mechanism and actuators as well as exceptional events from the middleware. It prevents actions that might immediately lead to destroying the platform, e.g. by mechanical collisions.

\section{Grand Challenges for Artificial Organism}
\label{sec:GrandChallenges}

Issues of challenges in evolutionary, reconfigurable and swarm robotics were mentioned several times since the early 1990s. We can refer to works \cite{Mataric96}, \cite{Ficici99}, \cite{Lipson00}, \cite{Sofge03}, \cite{Teo04} related to challenges with fitness estimation, ``reality gap" and others, whereas more recent work gives overview of challenges in the robotic area~\cite{Siciliano08}, such as over-motorization of reconfigurable systems or communication in swarm robotics. However, artificial organisms combine all three areas, resulting not only in a combination of problems and advantages, but also in qualitatively new challenges and breakthroughs. To demonstrate these breakthroughs, two Grand Challenges have been developed. The two following sections discuss underlying ideas of these Grand Challenges and problems in achieving them.

One of the important aspects of artificial organisms is their high degree of adaptivity. Moreover, adaptivity is estimated as one of the major technological challenges, see e.g.~\cite{Anderson05},~\cite{Astrom89},~\cite{Rohrs85}. On the other hand, one of the essential general challenges in robotics is a long-term independency of autonomous systems. It seems reasonable that Grand Challenges have to reflect these two issues.

\begin{figure}[htp]
\centering
\includegraphics[width=.48\textwidth]{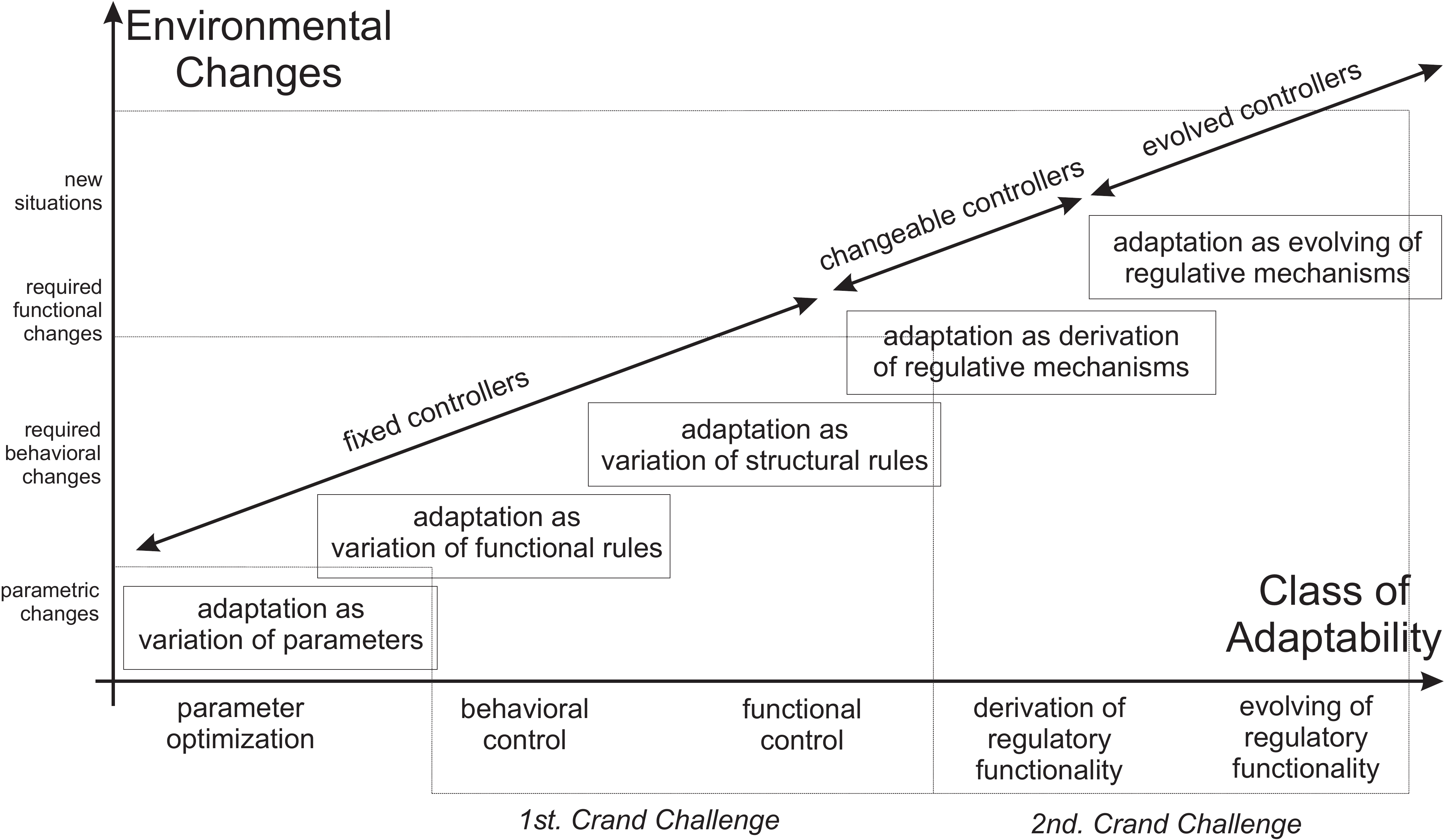}
\caption{Different adaptivity mechanisms in collective systems, from~\cite{Levi10}.
\label{fig:sec21-adaptiveMechanisms}}
\end{figure}

However, adaptivity is addressed by two Grand Challenges in different ways.
In Fig.~\ref{fig:sec21-adaptiveMechanisms} we represented a brief overview of different adaptive mechanisms, related to changes of environment (endogenous factors) and developmental plasticity of regulative mechanisms.
This figure can be roughly divided into low-, middle- and highly-rate adaptive parts (for regulative structures and corresponding environmental changes). Due to the nature of the Cognitive and Evolutionary frameworks, they address different adaptive parts: the 1st. Grand Challenge -- the medium-rate adaptive part and the 2nd. Grand Challenge -- the high-rate adaptive part.

\begin{figure}[htp]
\centering
\subfigure[]{\includegraphics[width=.25\textwidth]{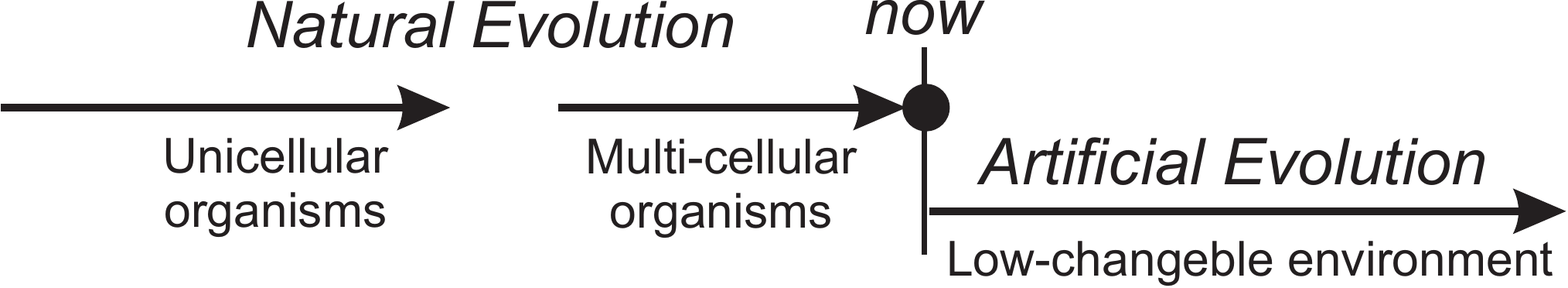}}~
\subfigure[]{\includegraphics[width=.25\textwidth]{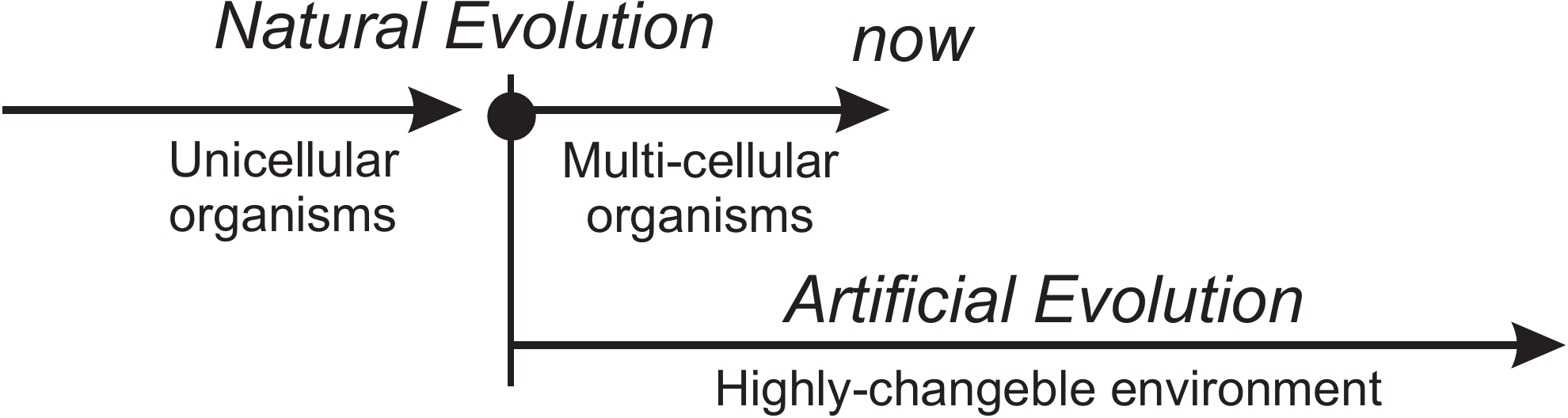}}
\caption{\textbf{(a)} Artificial evolution as a process following up natural evolution; \textbf{(b)} Artificial evolution as a process parallel to natural evolution. \label{fig:figCh}}
\end{figure}

Another split between Grand Challenges can be based on different understanding of artificial evolution. From the first viewpoint, artificial evolution is based on all achievements of natural evolution, including human technological progress, see Fig.~\ref{fig:figCh}(a). In other words, artificial evolution can be based on technological artefacts, pre-programmed behavioral patterns or include human-written algorithms. From another viewpoint, shown in Fig.~\ref{fig:figCh}(b), artificial evolution is considered as a process running parallel to natural evolution.
Arguments towards this position are very impressive achievements of natural evolution and attempt to understand and possibly to repeat them. Both viewpoints are interesting from philosophical, scientific and technological perspectives and can underlie both Grand Challenges.

Finally, due to the nature of the first Grand Challenge this should more strongly address the problems and advantages provided by cognitive approaches, whereas the second Grand Challenge should focus more on evolutionary ways of problem solving. It should be also mentioned that all Grand Challenges are envisaged and prepared as long-term goals, reflecting principal problems and breakthroughs. Their full realisation in the framework of academic research projects will be very challenging not least because of the numerous engineering problems.

\subsection{1st Grand Challenge -- 100 Robots, 100 Days}

The first Grand Challenge is primarily related to the Cognitive framework and addresses the problems of long-term independency in a medium-rate changeable environment with the assumption that artificial evolution can include technological artifacts. Here we can also find application and utilization of almost all other robotic issues such as e.g. reliability, energetic homeostasis, regulatory autonomy and others. This Grand Challenge may have the following form:

\emph{A large-scale system, let assume with 100 heterogeneous modules, is placed in a previously unknown area, which has complex, but structured character. This environment is slowly changing, for example, energetic resources are displaced or their indication is changing. This area contains enough energetic resources, such as power sockets or power cubes, which are sufficient for these 100 modules to survive in such an environment. The main energy source -- power sockets -- are inaccessible for individual robots, e.g. placed 30-40~cm above ground or in some structural gaps. Moreover, power sockets are switching on and off over the time in different order so that robots should first recognize position and quality of energy. Under these conditions the robots can survive only collectively, when aggregating into organisms with more distributed recognition and and extended affordance and actuation capabilities than individual robots. Aggregated robots perform in this area surveillance and disposal tasks with respect to fellow robots or modules passed away by pulling and carrying them if possible to a 'graveyard' - taking the environmental dynamics and the robots energy constraints into account. This experiment takes 100 days and should ideally be performed without any human maintenance work or supervision.}

\begin{figure}[htp]
\centering
\subfigure{\includegraphics[width=.45\textwidth]{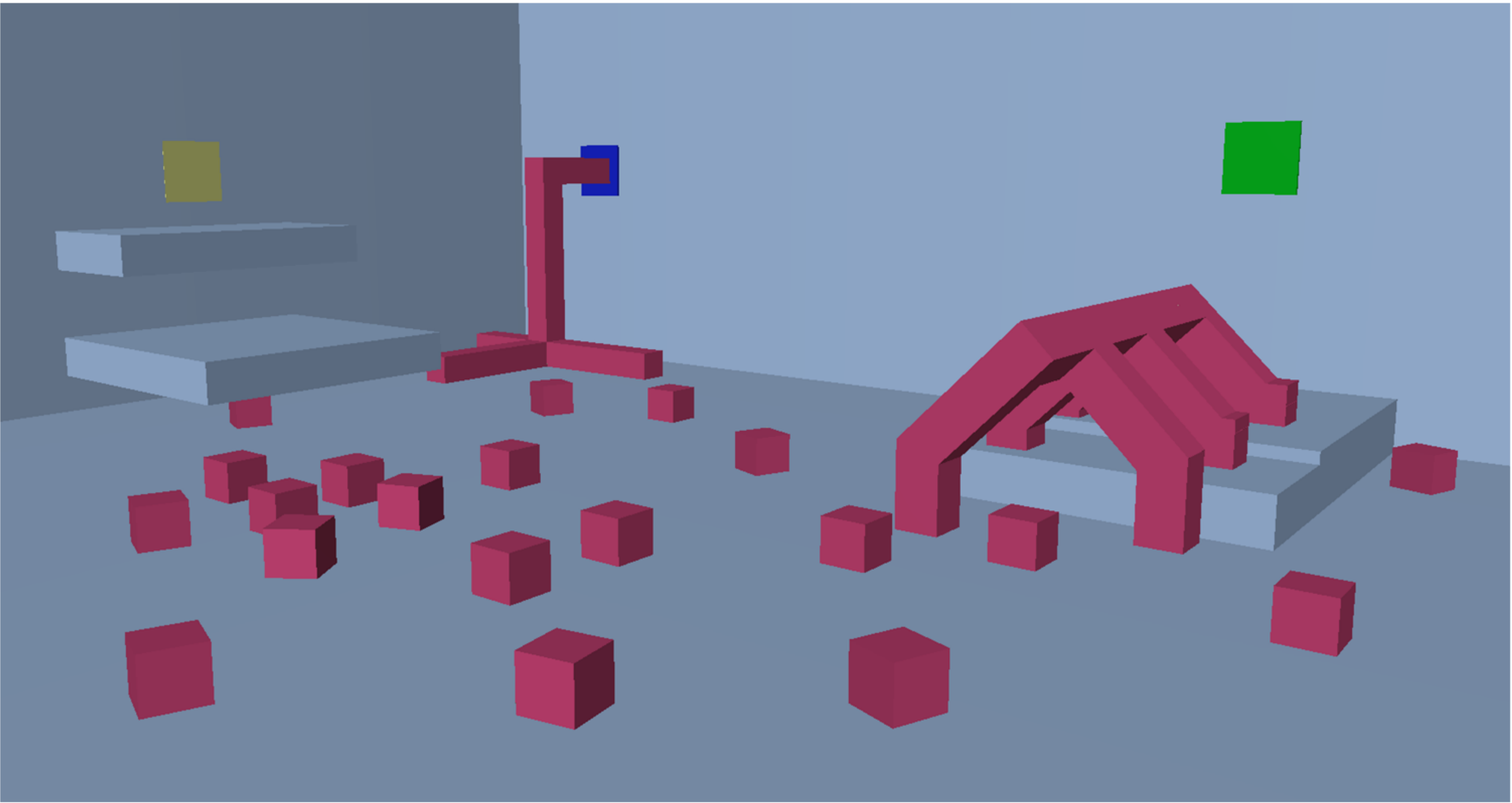}}
\caption{The sketch of the first Grand Challenge, colored boxes on the wall mean docking station (power sockets) - graveyard not depicted.\label{fig:fig4}}
\end{figure}

This idea is sketched in Fig.~\ref{fig:fig4}, different possible sub-scenarios and evaluation criteria are summarized in Table~\ref{tab:GC1}.

\begin{table}[ht]
\caption{Short overview of different possible sub-scenarios and evaluation criteria for the first Grand Challenge.\label{tab:GC1}}
\small
\begin{tabular}{p{.1cm}p{1.3cm}@{\extracolsep{3mm}}p{6.4cm}}\hline
N & Sub-scenarios & Comment \\\hline
1 &{Learning of environ-mental dynamics.} & After deployment in swarm or organism modes on a large area the robots that fail should not be a hazard and utilize remaining functions for common benefit. Furthermore, these robot modes should distill short-term survival strategies and long-term survival strategies. \\
2 & {Cognitive reconfigurability.} & Using different sensing/actuation and other cognitive capabilities of a swarm-organism mode to explore and to cope with given dynamic environmental-systemic conditions are a necessity for short-term survival.\\
3 & {Evaluating morpho-dynamic modes.} & Exploring and assessing fitness of structural and functional reconfigurations of diverse swarm-organism modes taking into account the dynamic environmental-systemic conditions are a necessity for long-term survival.\\
\hline
N & Evaluation crit. & Comment \\\hline
1 & Survived robots & Number of survived robots after N days\\
2 & Cognitive embodiment & Performance levels of morphodynamic pattern learning, recognition and generation (object recognition and avoidance); focus, selection and shifting of attention; situational awareness; anticipation / prediction by diverse swarm-organism modes under different dynmic environmental conditions.\\
3 & SW-HW Ratio & e.g. Number of energetically dead-robots (degree of adaptivity) compared to the hardware-dead robots .\\
\hline
\end{tabular}
\vspace{-7mm}
\end{table}

\section{Conclusion}
\label{sec:conclusion}

In this paper we presented the current development of the reconfigurable robotic platform which is capable of working as independent robot swarm as well as aggregated organisms. We have indicated three key capabilities of the platform: autonomous
morphogenesis, performing on-line and on-board evolving approaches and on-board fitness measurement. For these capabilities a mechatronic architecture and a Grand Challenge have been presented.

\section*{Acknowledgement}

The ``SYMBRION" project is funded by the European Commission
within the work programme ``Future and Emergent Technologies
Proactive" under the grant agreement no. 216342. The
``REPLICATOR" project is funded within the work programme
``Cognitive Systems, Interaction, Robotics" under the grant
agreement no. 216240. Additionally, we want to thank all
members of the projects for fruitful discussions.

%\bibliographystyle{unsrt}
%\bibliography{bibliography,bibliography_10}

\end{document}